\documentclass[runningheads]{llncs}

\usepackage[mobile]{eccv}

\usepackage{eccvabbrv}

\usepackage{graphicx}
\usepackage{booktabs}
\usepackage{wrapfig}
\usepackage{colortbl}
\usepackage{xcolor}    
\usepackage[accsupp]{axessibility}  

\usepackage{booktabs}
\usepackage{multirow}

\usepackage{hyperref}

\usepackage{orcidlink}

\begin{document}

\title{InterDyad: Interactive Dyadic Speech-to-Video Generation by Querying Intermediate Visual Guidance} 

\titlerunning{InterDyad}

\author{Dongwei Pan\inst{1} \and
Longwei Guo\inst{1} \and
Jiazhi Guan\inst{1} \and
Luying Huang\inst{1} \and
Yiding Li\inst{1} \and
Haojie Liu\inst{1} \and
Haocheng Feng\inst{1} \and
Wei He\inst{1} \and
Kaisiyuan Wang\inst{1}$^\dagger$ \and
Hang Zhou\inst{1}$^\dagger$
}

\authorrunning{Dongwei Pan, et al.}

\institute{Baidu Inc.\\
\email{\{pandongwei,wangkaisiyuan,zhouhang09\}@baidu.com} \thanks{$\dagger$ Corresponding authors}}

\maketitle

\begin{center}
 \centering
 \includegraphics[width=\textwidth]{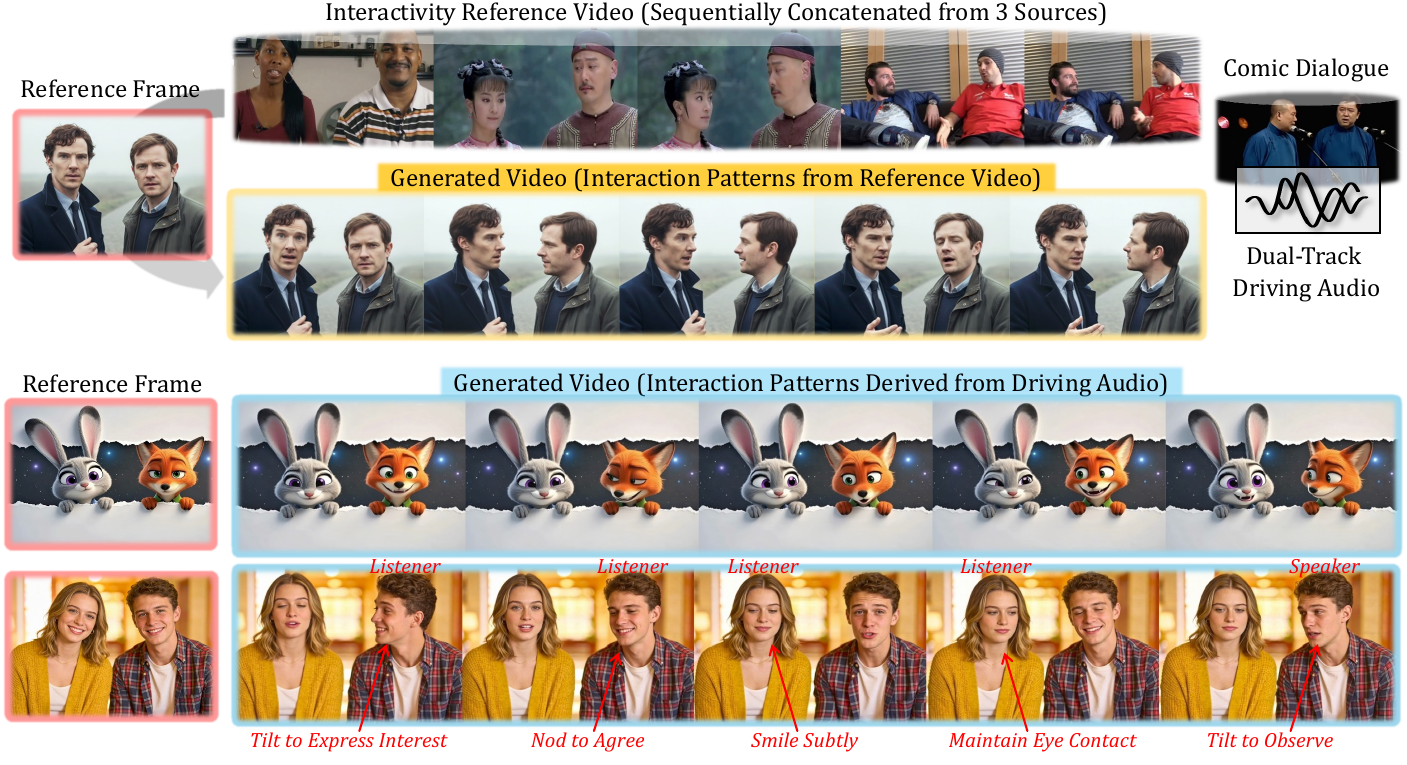}
\captionof{figure}{
\textbf{Dyadic-Conversational Video Generated by InterDyad.} 
Our method generates visual-audio synchronized conversational videos conditioned on a single reference frame of two subjects and dual-track driving audio, either replicating interaction patterns from a reference video ({\color{Goldenrod}yellow results}) or synthesizing plausible interactions directly from the driving audio ({\color{blue}blue results}).
}
\label{fig:teaser}
\end{center}

\begin{abstract}
Despite progress in speech-to-video synthesis, existing methods often struggle to capture cross-individual dependencies and provide fine-grained control over reactive behaviors in dyadic settings. To address these challenges, we propose InterDyad, a framework that enables naturalistic interactive dynamics synthesis via querying structural motion guidance. Specifically, we first design an Interactivity Injector that achieves video reenactment based on identity-agnostic motion priors extracted from reference videos. 
Building upon this, we introduce a MetaQuery-based modality alignment mechanism to bridge the gap between conversational audio and these motion priors. By leveraging a Multimodal Large Language Model (MLLM), our framework is able to distill linguistic intent from audio to dictate the precise timing and appropriateness of reactions. 
To further improve lip-sync quality under extreme head poses, we propose Role-aware Dyadic Gaussian Guidance (RoDG) for enhanced lip-synchronization and spatial consistency. Finally, we introduce a dedicated evaluation suite with novelly designed metrics to quantify dyadic interaction. Comprehensive experiments demonstrate that InterDyad significantly outperforms state-of-the-art methods in producing natural and contextually grounded two-person interactions. Please refer to our project page for demo videos: ~\url{https://interdyad.github.io/}.
\end{abstract}
\section{Introduction}

The rapid advancement of video generation foundation models~\cite{wan, hunyuanvideo, CogVideoX,fan2025vchitect} has led to significant milestones in producing high-fidelity single-person animations~\cite{wans2v, infinitetalk, infinityhuman, hunyuanvideo-avatar} from speech audio, revolutionizing applications including education, financial services, and digital entertainment.
This progress naturally extends toward multi-person co-presence scenarios, such as talk shows or situational comedies, where the challenge lies in synthesizing multiple interacting digital humans within a unified canvas to create cohesive multimedia experiences.

To extend generative capacity from individual to multiple subjects, several methods~\cite{multitalk, infinitetalk, bindyouravatar, anytalker, hunyuanvideo-avatar, omnihuman15, klingavatar2, tu2025stableavatar} have explored diverse audio-to-character binding strategies to achieve precise correspondence control.
Concurrently, large-scale multimodal generative models~\cite{ovi,mova,hacohen2026ltx,seedance15pro} have also demonstrated the ability to synthesize multi-person communication scenes by jointly generating audio and visual contents.
Despite these advancements, current paradigms primarily focus on synchronization between dual-stream audio and individual motion primitives (e.g., lip and limb dynamics), they largely overlook the underlying interpersonal correlations between acoustic cues and character interactions. Consequently, these models fail to capture the reactive non-verbal dynamics essential for realistic multi-person synthesis, leading to not coherent results.

Our objective is to develop a cost-effective dyadic human video generation framework that synthesizes two characters engaging in coherent, interactive behaviors with the driving speech. However, several challenges exist in achieving such a goal:
\textbf{1)} A straightforward idea is to drive dyadic interactions by manual text prompts. However, a fundamental discrepancy exists between static linguistic descriptions and the dynamic rhythmic nuances of speech. This reliance on high-level text inherently lacks the temporal precision required to trigger subtle reactive behaviors, such as a timely nod synchronized with a speaker's emphasis. While scaling datasets with dense linguistic annotations could theoretically enhance control, the requirement for such granular labeling presents a significant scalability bottleneck due to the prohibitive manual overhead.
\textbf{2)} Predicting interactive dynamics directly from speech audio is complicated by the inherent cross-modality discrepancy. The absence of an explicit bridge between a speaker’s audio and a listener’s response hinders the direct prediction of reactive movements, which consequently results in the difficulty of learning temporal synchrony essential for natural interaction.
\textbf{3)} Natural interaction necessitates frequent head rotations and mutual gaze, leading to large-angle profile views. Such extreme poses often result in the degradation of lip-sync accuracy and facial 
identity, as standard Speech-to-Video Generation backbones are typically optimized for near-frontal or limited-range motion.

To cope with the aforementioned issues, we present \textbf{InterDyad}, Interactive Dyadic Speech-to-Video Generation by Querying Intermediate Visual Guidance Framework.
Our key insight is \textit{to construct a bridge between speech audio and dyadic interactive dynamics by leveraging identity-agnostic motion representations derived from interactivity reference videos}. 

Specifically, we propose a two-stage training procedure designed to eliminate cross-modal discrepancy. In the first stage, we introduce a video reenactment branch named \textbf{Interactivity Injector} that characterizes interactive dynamics by extracting motion latents as primary driving signals for interactions. By integrating these visual-based motion priors into a pretrained audio-conditional video generation backbone, we achieve explicit dyadic pose control without compromising audio-visual synchronization. This stage ensures our model can prioritize high-fidelity, natural motion synthesis independently of the complexities involved in direct audio-to-video mapping.

In the second stage, we leverage MetaQuery~\cite{pan2025transfermodalitiesmetaqueries} as \textbf{Modality Alignment} to bridge the remaining gap between the speaker’s audio and the pre-established motion space. By using MetaQuery as a learnable interface, we effectively map the speech signals into the motion space optimized in the first stage. Specifically, we employ a Multimodal Large Language Model (MLLM) to distill linguistic intent from the speaker's audio. This design ensures that reactive behaviors—such as responsive postural shifts—are synthesized with superior temporal synchrony and contextual groundedness.
To maintain high-fidelity synthesis under the resulting extreme perspectives, we propose Role-aware Dyadic Gaussian Guidance (RoDG), which implements a controllable Gaussian distribution to adaptively prioritize audio-conditioned CFG within the oral region. By imposing a spatially-weighted emphasis on acoustic-visual synchronization, this strategy ensures robust lip-sync accuracy even during significant postural transitions.

Our contributions are summarized as:
\textbf{1)} We propose InterDyad, a Dyadic-Conversational Generation Framework that enables explicit orchestration of interaction patterns through the Interactivity Injector and the MetaQuery-based Modality Alignment mechanism. By leveraging MetaQueries to translate conversational audio into interaction priors, this design effectively bridges the modality gap and enables characters to exhibit contextually grounded non-verbal reactions.
\textbf{2)} We optimize dyadic interaction quality via Role-aware Dyadic Gaussian Guidance (RoDG), which adaptively intensifies speaker-specific acoustic constraints to ensure precise lip-synchronization and reactive coordination.
\textbf{3)} We design a dedicated evaluation suite for dyadic interactivity, introducing metrics to quantify interaction in spatio \& temporal coordination, with extensive experiments validating our framework's superiority.
\section{Related Work}

\subsection{Audio-driven Video Generation for Multi-person Scenes}

Speech-driven human video generation has gradually shifted from two-stage motion–rendering pipelines to diffusion-based end-to-end synthesis conditioned on audio and appearance~\cite{guo2024liveportrait,wei2024aniportrait,guan2025audcast,tian2024emo,wans2v,hunyuanvideo-avatar,omniavatar,omnihuman1}.
Recent methods extend this paradigm to multi-person scenarios by exploring different strategies for audio-to-character binding and spatial conditioning.
For multi-speaker S2V, \textit{MultiTalk}~\cite{multitalk} introduces Label Rotary Position Embedding (L-RoPE) to associate identities with their corresponding audio streams. 
Building upon this, \textit{InfiniteTalk}~\cite{infinitetalk} proposes a sparse-frame reference mechanism to address consistency and enable coherent video dubbing for long-duration sequences. 
Subsequently, \textit{LongCat-Video-Avatar}~\cite{longcat} ensures natural body behaviors by disentangling audio signals from motion dynamics during extended synthesis.
Additionally, some other works ~\cite{anytalker,omnihuman1,interacthuman} enhance robustness in complex identity binding and large-scale animation through identity-aware attention, diverse motion conditions, or layout-aware masks.
In parallel with the evolution toward Unified Audio-Visual Generation, multimodal generative models such as \textit{MoVA}~\cite{mova} and related unified audio-visual generators~\cite{ovi,seedance15pro,hacohen2026ltx,guan2024resyncer,guan2023stylesync} further demonstrate the ability to synthesize synchronized multi-person audiovisual content through joint token generation.
However, these visual/audio-visual generators primarily emphasize synchronization and identity alignment, lacking explicit designs for complex interactive dynamics. In contrast, our framework enables direct cross-modal steering for dyadic interaction patterns through an Interactivity Injector and a MetaQuery-based Modality Alignment mechanism, ensuring both interactivity and precise coordination.

\subsection{Interactivity Control and Listener-aware Generation}
Listener-aware generation aims to synthesize non-verbal feedback conditioned on speaker signals~\cite{responsive}. To this end, \textit{INFP}~\cite{infp} adopts a two-stage design that maps dual-track audio to a learned motion latent space, enabling automatic transitions between speaking and listening states; \textit{DiTaiListener}~\cite{ditailistener} leverages video diffusion with a segment-based design to enhance listener response consistency conditioned on the speaker's facial motions. 
Beyond offline synthesis, empowered by recent advancements in autoregressive video diffusion that tackle error accumulation for minute-scale and real-time generation~\cite{chen2024diffusion, cui2025self, huang2025self, liu2025rolling}, recent works have explored streaming conversational avatars for real-time interaction~\cite{infinitetalk, flowact, ki2026avatar}. Specifically, \textit{Live Avatar}~\cite{liveavatar} enables prompt-driven behavior transitions under strict latency budgets by optimizing inference pipelines. 
However, most existing conversation synthesis systems~\cite{from_audio_to_photoreal, lets_go_real_talk, dialoguenerf} are restricted to split-screen or frontal settings, which only account for individual behaviors in isolation. While recent advancements have made strides in capturing holistic human motions—such as \textit{GestureHYDRA}~\cite{yang2025gesturehydra}, which utilizes a hybrid modality Diffusion Transformer and retrieval-augmented generation for semantic co-speech gesture synthesis—achieving explicit interaction guidance and modeling the joint interaction between speaker and listener in shared-canvas video scenes remains formidable. Motivated by these challenges, InterDyad introduces a unified framework integrating reference-driven interaction control with MLLM-based reasoning for coherent dyadic video synthesis.
\section{Method}

Given a visual reference image $I$ containing two subjects denoted by $k \in \{1, 2\}$, dual-track audio streams $\mathbf{a} = \{\mathbf{a}_1, \mathbf{a}_2\}$, and a text prompt $c_\text{text}$, our primary objective is to synthesize a sequence of dyadic video frames $\mathbf{V} = \{v_1, v_2, \dots, v_T\}$. The generated video must ensure that both characters maintain strict identity consistency, precise audio-lip synchronization, and contextually appropriate inter-agent dynamics.
As illustrated in Fig.~\ref{fig:pipe}, we propose InterDyad, a \textbf{Dyadic-Conversational Generation Framework} designed for end-to-end video synthesis conditioned on dual-person dialogue input. 

\subsection{Preliminaries}

\subsubsection{Video Foundation Model.}
We build our framework upon a flow-based Diffusion Transformer (DiT)~\cite{wan}. During training, given a visual reference $I$ and a target video $\mathbf{V}$, a VAE~\cite{wan} encodes them into a spatio-temporal latent $\mathbf{z}_0 \in \mathbb{R}^{C \times T \times H \times W}$. Conditions including text $c_\text{text}$, image $c_\text{img}$ and audio embeddings $c_\text{audio}$ are formulated as a joint condition set $\mathbf{c} = \{c_\text{text}, c_\text{img}, c_\text{audio}\}$. The DiT learns a target velocity field $\mathbf{v}_t$ by minimizing the flow matching objective:
\begin{equation}
\mathcal{L}_\text{LDM} = \mathbb{E}_{\mathbf{z}_0, \mathbf{z}_1, t} \left[ \| \mathbf{v}_t(\mathbf{z}_t, t, \mathbf{c}) - (\mathbf{z}_0 - \mathbf{z}_1) \|_2^2 \right],
\end{equation}
where $\mathbf{z}_t = t\mathbf{z}_0 + (1-t)\mathbf{z}_1$ interpolates between the data $\mathbf{z}_0$ and initial noise $\mathbf{z}_1 \sim \mathcal{N}(0, \mathbf{I})$ at timestep $t \sim \mathcal{U}(0,1)$. For long-duration generation, we employ a multi-clip inference strategy, utilizing motion latents as temporal anchors and $I$ as a global anchor to ensure temporal consistency and identity preservation.

\subsubsection{Speech-Driven Two-Person Generation.}
Generating dyadic videos requires processed dual-track audio streams ($\mathbf{a}_1, \mathbf{a}_2$). Following InfiniteTalk~\cite{infinitetalk}, we adopt Audio Cross-Attention equipped with Label Rotary Position Embedding (L-RoPE)~\cite{multitalk} for precise audio-spatial alignment. L-RoPE modulates the queries $\mathbf{q}_m$ and keys $\mathbf{k}_n$ using identity-specific spatial labels $l$:
\begin{equation}
\hat{\mathbf{q}}_m = \mathbf{q}_m e^{i l_m \theta_\text{base}}, \quad \hat{\mathbf{k}}_n = \mathbf{k}_n e^{i l_n \theta_\text{base}},
\end{equation}
where $i$ is the imaginary unit, $m, n$ denote token indices, and $\theta_\text{base}$ is a base angle. By assigning distinct label ranges to the visual tokens of Person 1 and Person 2 via spatial masks, and matching them with corresponding audio embeddings, this mechanism strictly binds acoustic features to their respective spatial latents.

\begin{figure*}[!t]
\centering
\includegraphics[width=\linewidth]{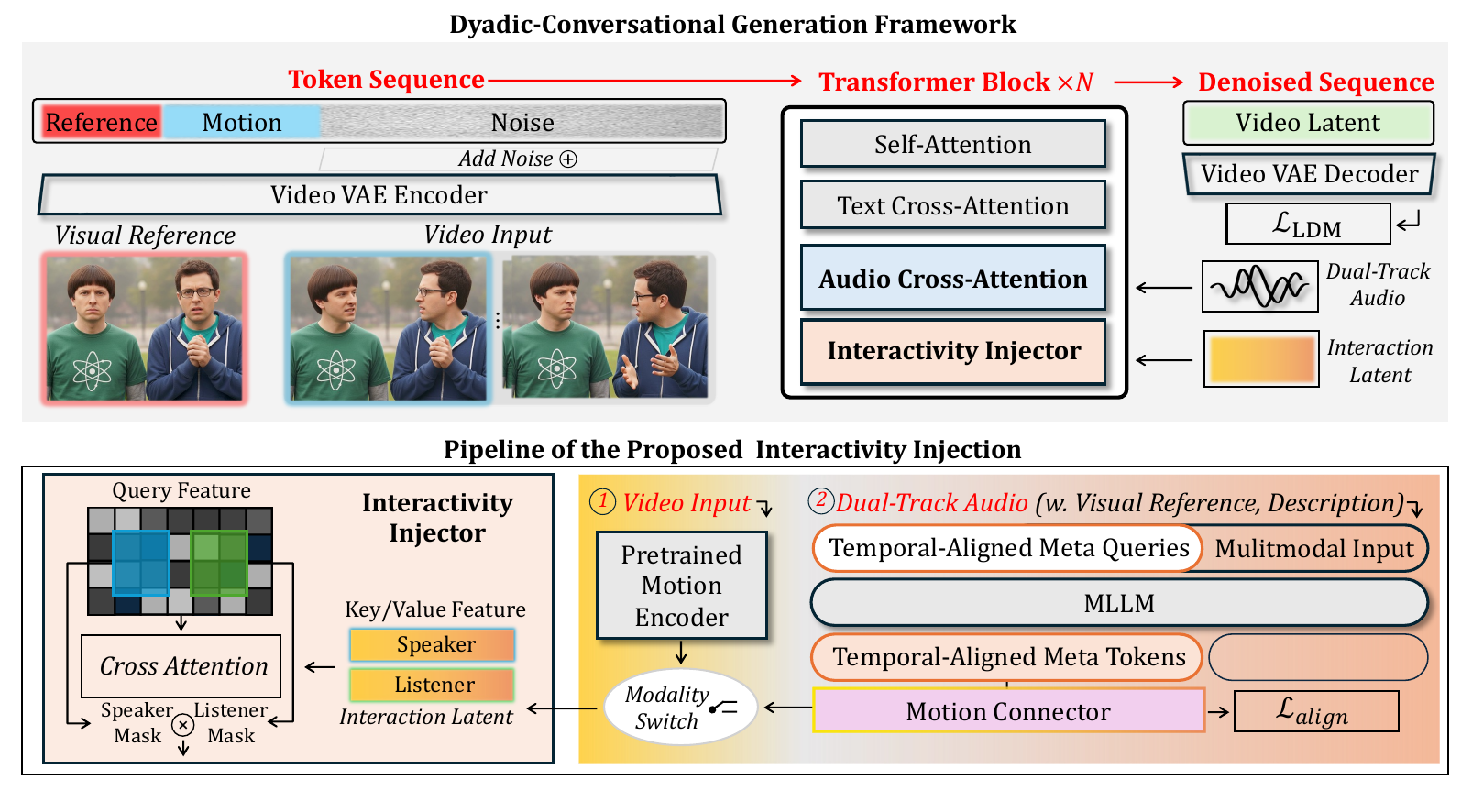}
\vspace{-0.25cm}
\caption{
\textbf{Overview of our InterDyad framework.}
\textbf{Top}: The overall architecture is constructed by sequentially stacking Transformer blocks and trained via denoising to synthesize dyadic conversational videos with synchronized audio-visual human dynamics and coherent inter-subject interactions.
\textbf{Bottom}: We illustrate the interactivity injection mechanism, which leverages switchable multimodal inputs to enable rich and controllable synthesis of interactive motion patterns.
}
\label{fig:pipe}
\vspace{-0.25cm}
\end{figure*}

\subsection{Dyadic-Conversational Generation Framework}

Current speech-to-video paradigms typically treat dyadic interaction as an implicit byproduct of joint generation, which often yields ``passive'' characters lacking fine-grained reactive behaviors. 
To bridge the gap between static co-presence and authentic interaction, we propose a framework that enables explicit orchestration of interaction patterns. 
Central to our approach is the formal definition of conversational roles within the shared canvas: for each interaction cycle, we assign the participants as the Speaker ($S$) and the Listener ($L$). This role-based formulation allows us to explicitly model the cross-individual dependency where the listener’s non-verbal feedback is dynamically conditioned on the speaker's cues.
As illustrated in the bottom part of Fig.~\ref{fig:pipe}, our framework operates through a dual-mechanism design tailored to the core challenges of dyadic synthesis.
To address the challenge of modeling stochastic cross-individual dependencies, we first introduce \textbf{Interactivity Injector} for the direct injection of behavioral priors. By treating interaction as a steerable modality, this module extracts \textit{Interaction Patterns} from \textit{Video Inputs} to prescribe specific states and subtle reactive behaviors (e.g., postural shifts or head nods) to either participant.
To inherently support direct, cross-modal guidance from conversational audio while bridging modality discrepancies, we further incorporate a \textbf{Modality Alignment} mechanism that leverages an intermediary \textbf{MetaQuery} to derive context-aware interaction patterns from the {\textit{Dual-Track Audio}}, fostering more natural and frequent interactivity between Speaker and Listener.

\subsubsection{Interactivity Injector.}
To achieve explicit guidance over dyadic interactions, we introduce the Interactivity Injector, which enables the framework to steer non-verbal speaker-listener behaviors directly via video inputs as visual reference.
Unlike conventional dyadic generation~\cite{multitalk,infinitetalk,longcat,wans2v}, where interaction states are often treated as implicit latent features entangled within the synthesis process, our module extracts high-level interaction patterns from a reference video input and injects them as prior-driven modulations into the synthesis process.

Following LIA~\cite{LIA,wang2024lia} and Wan-Animate~\cite{cheng2025wananimateunifiedcharacteranimation}, we employ a pre-trained motion encoder to extract identity-agnostic interaction latent $\mathbf{m}_k$ for  Speaker and Listener $k \in \{S, L\}$ from reference videos. This ensures the captured dynamics are strictly decoupled from appearance, providing steerable interaction priors for the synthesis.
To avoid modal conflicts between reference lip movements and target audio, we apply a \textit{Lips-Masking Strategy}: the oral regions are masked before encoding. This ensures the extracted priors—such as head tilts and nods—focus exclusively on non-verbal dynamics, preventing interference with audio-driven lip synchronization.

To integrate the extracted interaction latent of both Speaker and Listener into the diffusion backbone while maintaining strict identity binding, we implement a straightforward yet effective \textit{Spatial-Masking Cross-Attention Mechanism}. This operation ensures that the behavioral cues for each participant are localized to their respective spatial coordinates, preventing feature leakage between the Speaker and Listener.
Specifically, we adaptively track the spatial localization of each individual to generate binary masks $\mathbf{M}_k \in \{0, 1\}$ for both the Speaker and Listener $k \in \{S, L\}$. These masks act as spatial constraints within the attention layers. The injection is formulated as:
\begin{equation}
\mathbf{z}_\text{inter} = \sum_{k \in {S, L}} \mathbf{M}_k \odot \text{Softmax}\left(\frac{\mathbf{Q} (\mathbf{K}_k)^\top}{\sqrt{d}}\right) \mathbf{V}_k,
\end{equation}
where $\mathbf{Q}$ is derived from the video latents $\mathbf{z}_t$, and $\mathbf{K}_k, \mathbf{V}_k$ are projected from the interaction latent $\mathbf{m}_k$ of person $k$. By restricting the attention influence within $\mathbf{M}_k$, we force a direct correspondence between reference dynamics and target identities. This mechanism allows for fine-grained control by enabling the injection of specific interaction patterns from a reference video during inference.

\subsubsection{Modality Alignment with MetaQueries.}
While the injection module enables explicit steering, synthesizing contextually appropriate interaction states remains a significant challenge due to the stochastic \textit{many-to-many mapping} between conversational audio and non-verbal interaction of dyadic speakers. 
To address this, we introduce a Modality Alignment module to translate acoustic contexts into synchronized motion dynamics as intermediate guidance, enabling to derive interaction patterns from dual-track audio.

Specifically, we leverage a frozen pre-trained MLLM (Qwen3-Omni~\cite{qwen3omni}) as a multimodal encoder to extract context-aware interaction priors. Beyond the dual-track conversational audio $a_{k}$, we provide the visual reference $I$ and text description $\mathbf{c}_\text{text}$ as additional inputs to the encoder. This allows the model to perceive the initial spatial relationship and identity-specific traits(Speaker/Listener) from the first frame, while grounding the interactive intent in the provided linguistic context.
To bridge the gap between these high-level multimodal cues and interaction patterns, we adopt the Meta-Queries~\cite{pan2025transfermodalitiesmetaqueries} paradigm by introducing a set of learnable query tokens. Unlike the original formulation that utilizes a unordered query set, we structure these tokens as a Temporal-Aligned Meta-Query Sequence $\mathcal{Q} \in \mathbb{R}^{N \times D}$. In our implementation, we enforce a strict one-to-one correspondence between the $N$ query tokens and the $N$ acoustic frames within each training crop, ensuring that each query token is anchored to a specific temporal position.
The interaction-specific hidden states, sliced from the MLLM backbone's output at the meta-query positions,  are subsequently refined by a \textit{Temporal Connector Network} comprising: (i) 1D-Convolutional layers for local neighbor-frame modeling; (ii) Transformer Encoder layers for long-range dependency capture; and (iii) a Linear layer for dimension modulation.
The connector outputs the predicted dyadic interaction patterns $\hat{\mathbf{m}}$. To ensure both motion fidelity and kinematic smoothness, we optimize the modality alignment module via a joint objective function:
\begin{equation}
\mathcal{L}_\text{align} = \|\mathbf{m} - \hat{\mathbf{m}}\|_2^2 + \|\Delta \mathbf{m} - \Delta \hat{\mathbf{m}}\|_2^2,
\end{equation}
where $\mathbf{m}$ represents the ground-truth interaction patterns, and $\Delta$ denotes the temporal difference operator. The first term, \textit{MSE loss} ensures point-wise alignment, while the second term, the \textit{temporal loss}, enforces temporal coherence and mitigates motion jitter. 
By optimizing this cross-modality mapping, our framework achieves explicit guidance over dyadic interactions directly from dual-track audio inputs, effectively deriving high-fidelity interactive patterns to steer the subsequent diffusion process.

\subsection{Dyadic Conversational Inference}

\subsubsection{Role-Aware Dyadic Gaussian Guidance (RoDG).}
A distinct difficulty in dyadic synthesis arises from frequent head rotations and mutual gaze, which lead to extreme profile views. Under these severe perspective shifts, the visual integrity of the oral region is often compromised, causing standard generation backbones to struggle with lip-synchronization accuracy.

To address this, we extend the dynamic guidance paradigm from OmniSync~\cite{peng2025omnisync} and propose Role-aware Dyadic Gaussian Guidance, which implements a role-adaptive optimization strategy during inference (as visualized in Fig.~\ref{fig:method_RoDG}). Specifically, during each denoising step, we detect the active Speaker $S$ and the Listener $L$ based on VAD-estimated speech activity from the two audio tracks, which is also used to derive a frame-wise boosting intensity curve $\boldsymbol{\alpha}=\{\alpha_t\}_{t=1}^{T}$. We then spatially localize the audio guidance by constructing identity-specific 2D Gaussian maps $\mathbf{G}_k$ for each person $k \in \{1, 2\}$. The speaker’s map is activated with a peak centered at the lip location $\mu_{\text{lip},S}$, while the listener’s map is set to zero to prevent audio guidance from leaking to the non-speaking mouth region. The audio Classifier-Free Guidance (CFG) scale $w_a$ is then spatially modulated as:
\begin{equation}
w_a(\mathbf{x},t) = w_\text{base} + \alpha_t \cdot \sum_{k\in{1,2}} \mathbb{I}(k=S_t)\cdot \mathbf{G}_k(\mathbf{x};\mu_{\text{lip},k},\sigma),
\end{equation}
where $\mathbb{I}(\cdot)$ is an indicator function for speaker activation, $w_\text{base}$ is the baseline scale, $\sigma$ controls the refinement radius, and $\alpha_t$ is the VAD-derived frame-wise boosting intensity at frame $t$. By concentrating the intensified acoustic conditioning precisely on the speaker's oral region while bypassing the listener, this mechanism ensures robust lip reconstruction from audio cues even under degraded visual signals. 

\subsection{Dyadic Interactivity Evaluation}

To quantitatively evaluate the nuanced coordination and mutual influence within two-person conversational scenes, we propose a comprehensive evaluation suite, specifically designed to capture dyadic interactivity across two complementary dimensions: temporal synchrony and spatial vitality.

\subsubsection{Dyadic Interaction Temporal Synchrony (DI-Sync).}
\begin{figure*}[htbp]
  \centering
  \begin{minipage}{0.45\textwidth}
    \centering
    \includegraphics[width=\textwidth]{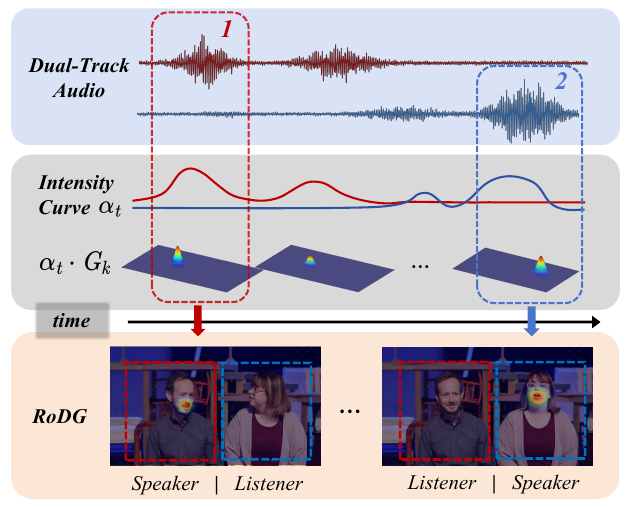}
    \caption{\textbf{RoDG.} RoDG uses dual-track VAD to assign Speaker and Listener over time, boosting audio guidance on the speaker's lip Gaussian while suppressing the listener to avoid cross-talk.}
    \label{fig:method_RoDG}
  \end{minipage}
  \hfill 
  \begin{minipage}{0.53\textwidth}
    \centering
    \includegraphics[width=\textwidth]{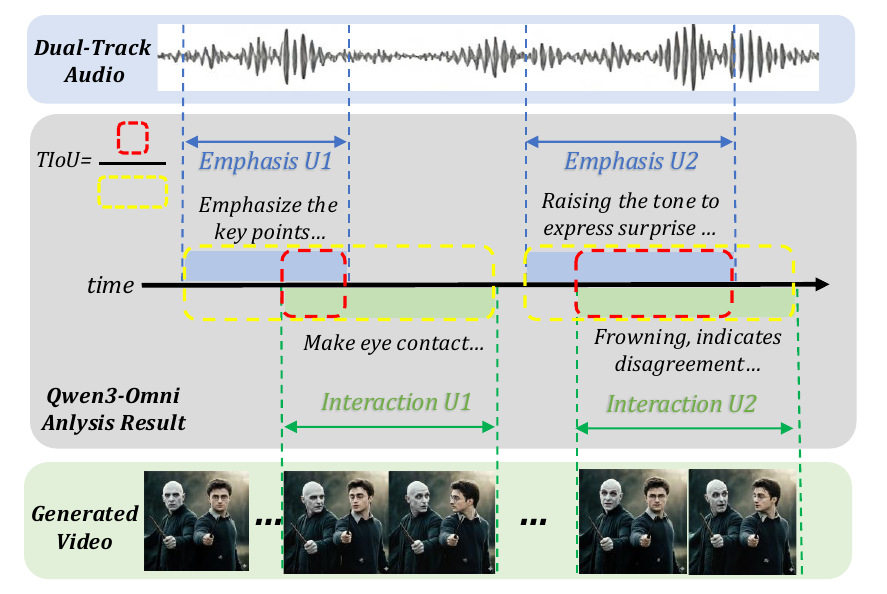}
    \caption{\textbf{DI-Sync.} Audio union of prosodic emphasis and Video union of reactive behaviors are extracted by MLLM, then Temporal {\color{red}{\textbf{Intersection}}} of {\color{yellow}{\textbf{Union}}}(TIoU) are calculated.}
    \label{fig:eval_metric}
  \end{minipage}
\end{figure*}

The \textbf{DI-Sync}(Fig. ~\ref{fig:eval_metric}) metric quantifies the causal coordination between the speaker’s verbal cues and the listener’s non-verbal responses. Unlike frame-level lip-sync metrics, DI-Sync focuses on high-level interaction timing. We leverage a Multimodal Large Language Model (Qwen3-Omni~\cite{qwen3omni}) to extract the temporal segments of prosodic emphasis from the audio ($\mathcal{T}_\text{audio}$) and the corresponding reactive behaviors from the generated video ($\mathcal{T}_\text{video}$). DI-Sync is formulated as the Temporal Intersection over Union (TIoU) with a social latency compensation $\delta$:

\begin{equation}
\text{DI-Sync} = \frac{|\mathcal{T}_\text{audio} \cap \text{Shift}(\mathcal{T}_\text{video}, \delta)|}{|\mathcal{T}_\text{audio} \cup \text{Shift}(\mathcal{T}_\text{video}, \delta)|}
\end{equation}

where $\text{Shift}(\cdot, \delta)$ aligns the listener's response by compensating for the natural human reaction delay. Following the empirical observation of social latency, $\delta$ is set to $0.5s$ in our experiments. A higher DI-Sync indicates superior temporal intelligence in responding to conversational stimuli.

\subsubsection{Dyadic Interaction Saliency (DI-Sali).}
Extending AnyTalker’s~\cite{anytalker} listener-centric intensity metric, we propose DI-Sali to evaluate the holistic engagement of dyadic scenes. Unlike isolated listener metrics, DI-Sali captures co-motion dynamics by aggregating inter-frame landmark displacements from both Speaker ($S$) and Listener ($L$), specifically targeting the signal-rich ocular and periorbital regions:

\begin{equation}
\text{DI-Sali} = \frac{1}{T-1} \sum_{t=1}^{T-1} \left(\Delta E_S(t, t+1) + \Delta E_L(t, t+1) \right)
\end{equation}

where $\Delta E$ denotes the mean Euclidean distance of eye-related landmarks. By measuring the interactive pair's collective movement, DI-Sali effectively quantifies the spatial vitality and mutual involvement of the generated conversation.

\section{Experiments}

\subsection{Datasets}
We establish a large-scale, high-fidelity dyadic interaction dataset through a \textbf{cascaded curation pipeline}. From an extensive pool of in-the-wild videos, we first apply heuristic thresholds to filter out clips with low resolution and severe camera jitter. Next, we employ DWPose~\cite{yang2023effective} to robustly isolate 4M dyadic candidates. To build strictly aligned audio-visual pairs, we perform synchronization processing to retain the full dyadic video sequences alongside the isolated acoustic tracks of each individual. Crucially, to ensure the dataset captures rich inter-agent dynamics rather than static talking heads, we enforce a subsequent DWPose-driven large-motion filter, refining the pool to 3M clips. After a final clarity check, we obtain 700,000 high-quality videos, all standardized to at least 720P, 25 FPS, and 3-10 seconds in length. A test set of \textit{100 diverse cases} is randomly sampled for evaluation. Further details are provided in the Supplementary Material.
\subsection{Experimental Settings}
\subsubsection{Implementation Details.}
We adopt Wan2.1-I2V-14B~\cite{wan} as our foundational video diffusion model. Following the generation paradigm of InfiniteTalk~\cite{infinitetalk}, we utilize Wav2Vec~\cite{baevski2020wav2vec20frameworkselfsupervised} for audio embedding and generate 81-frame video chunks conditioned on a context window of 9 images. To leverage robust generative priors, we use the pretrained motion encoder from Wan-Animate~\cite{cheng2025wananimateunifiedcharacteranimation}, and initialize the cross-attention layer in Interactivity Injector also from Wan-Animate. The entire framework is optimized through a streamlined two-stage training paradigm:

\textbf{Stage 1: Modality Alignment.} We exclusively train the Modality Alignment module to bridge the gap between audio modal and interaction patterns. Utilizing the temporal-aligned Meta-Queries~\cite{pan2025transfermodalitiesmetaqueries}, this stage distills context-aware interaction priors from the frozen Qwen3-Omni~\cite{qwen3omni} encoder. This phase establishes a robust mapping from conversational cues to motion space, converging in $12$ hours with $6,000$ iterations.

\textbf{Stage 2: End-to-End DiT Fine-Tuning.} We jointly optimize the Interactivity Injector and the Audio Cross-Attention. This stage forces the model to integrate interaction patterns with precise identity binding via the Identity-Aware Cross-Attention mechanism. This end-to-end process requires approximately 2 days to complete $10,000$ iterations.

\subsubsection{Comparison methods.}
We compare with representative speech-to-video generation methods, including \textbf{MultiTalk}\cite{multitalk}, \textbf{InfiniteTalk}\cite{infinitetalk}, and \textbf{LongCat-Video-Avatar}\cite{longcat}.
For each baseline, we follow the official inference settings and use the same prompts and evaluation protocol whenever applicable.
\subsubsection{Evaluation Metrics.}
We comprehensively assess our framework across three core dimensions:

\textbf{Visual \& Identity Quality}: We report Fréchet Video Distance (FVD)~\cite{unterthiner2018towards} and Fréchet Inception Distance (FID)~\cite{heusel2017gans} to evaluate the visual fidelity of the generated frames. To measure Identity Consistency (ID-Cons) in dyadic scenarios, we first extract the highest-confidence face regions for both individuals from the ground-truth video as references. We then compute the ArcFace~\cite{Deng_2022} similarity score between these reference crops and the faces detected in the generated frames, verifying the strict preservation of the subjects' facial appearances.

\textbf{Audio-Visual Synchronization}: We utilize Sync-C, Sync-D scores to rigorously measure the accuracy of audio-driven lip movements.

\textbf{Interactivity Reciprocity (Ours)}: To measure the interaction dynamics between speaker and listener, we propose two new metrics. Dyadic Interaction Temporal Synchrony (DI-Sync) measures the causal timing alignment between the active speaker's audio cues and the listener's non-verbal reactions. Dyadic Interaction Saliency (DI-Sali) evaluates the co-motion dynamics by integrating inter-frame displacements from both the Speaker and the Listener.

\subsection{Evaluation Results}
\subsubsection{Quantitative Comparison.}
We evaluate our method against state-of-the-art baselines on our curated dyadic conversational dataset. As shown in Tab.~\ref{tab:main_results}, our approach demonstrates competitive or better performance across most dimensions. In terms of visual fidelity, our method improves upon existing frameworks, yielding better scores in both FID and FVD. While InfiniteTalk achieves a slightly higher ID-Cons score, this difference can be attributed to the enhanced interactivity produced by our framework. Specifically, our method generates more natural and dynamic behaviors, frequently resulting in challenging poses such as side-profile views, which inherently complicate face recognition and identity preservation metrics. This increased level of dynamic engagement is quantitatively supported by our higher scores on the proposed interactivity metrics. Furthermore, generating accurate lip movements in dyadic scenarios is challenging due to face-to-face spatial orientations. Nevertheless, our framework shows strong audio-visual alignment, achieving better Sync-C and Sync-D scores~\cite{chung2016out} than the baselines.

Regarding the proposed interactivity metrics, our method shows clear improvements. Because existing baselines typically lack dedicated interaction modules, they often produce static or less responsive listener behaviors. By explicitly modeling interaction dynamics via the Modality Alignment mechanism, our approach attains higher DI-Sync and DI-Sali scores compared to the evaluated baselines. This supports the effectiveness of our framework in generating more natural, temporally synchronized, and contextually appropriate reciprocal reactions in two-person conversations.

\begin{table}[t]
  \centering
  \caption{Quantitative comparison on our curated dyadic conversational dataset. LongCat-VA denotes LongCat-Video-Avatar. ``$\uparrow$'' indicates higher is better, and ``$\downarrow$'' indicates lower is better. Best results are \textbf{bolded}, and the second best are \underline{underlined}.}
  \label{tab:main_results}
  \resizebox{0.85\linewidth}{!}{
  \begin{tabular}{c ccc cc cc}
    \toprule
    \multirow{2}{*}{Method} & \multicolumn{3}{c}{Visual \& Identity} & \multicolumn{2}{c}{Synchronization} & \multicolumn{2}{c}{Interactivity (Ours)} \\
    \cmidrule(lr){2-4} \cmidrule(lr){5-6} \cmidrule(lr){7-8}
    & FID $\downarrow$ & FVD $\downarrow$ & ID-Cons $\uparrow$ & Sync-C $\uparrow$ & Sync-D $\downarrow$ & DI-Sync $\uparrow$ & DI-Sali $\uparrow$ \\
    \midrule
    MultiTalk \cite{multitalk}       & 49.6047 & \underline{477.6189} & 0.5275 & \underline{3.2253} & 5.8941 & 0.2333 & 0.8889 \\
    InfiniteTalk \cite{infinitetalk} & 46.3762 & 440.6732 & \textbf{0.6418} & 3.0446 & \underline{5.5412} & 0.2371 & 0.8560 \\
    LongCat-VA \cite{longcat}        & \underline{45.4698} & 548.8332 & 0.6059 & 3.1985 & 5.7200 & \underline{0.2417} & \underline{1.1145} \\
    \midrule
    \textbf{Ours}                    & \textbf{38.3260} & \textbf{415.5064} & \underline{0.6310} & \textbf{3.3067} & \textbf{4.1786} & \textbf{0.2747} & \textbf{1.2349} \\
    \bottomrule
  \end{tabular}
  }
\end{table}

\subsubsection{Qualitative Comparison.}
\begin{figure}
    \centering
    \includegraphics[width=\linewidth]{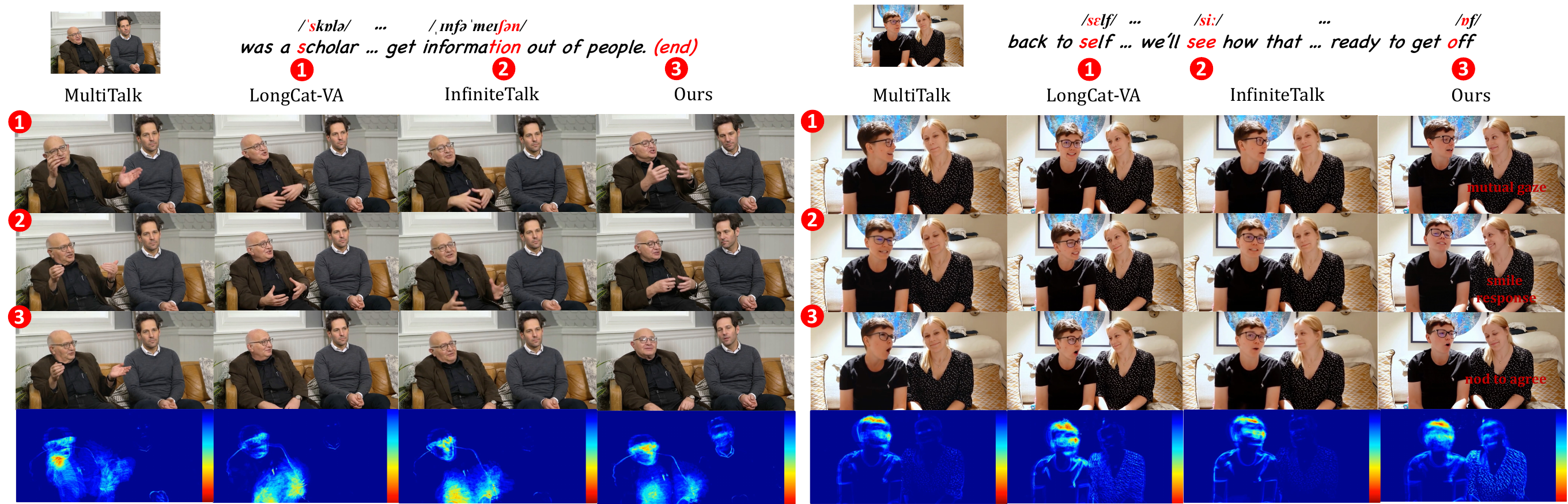}
    \caption{\textbf{Qualitative comparison with existing baselines.} 
    The figure illustrates the generated inter-subject dynamics from different methods, where the numbers in red dots indicate three different timestep.
    Accumulated motion heatmaps are provided in the bottom row to visualize reaction intensity.
    }
    \label{fig:exp_main}
\end{figure}

To further evaluate visual fidelity and inter-subject dynamics, we present qualitative comparisons between InterDyad and existing baselines, as illustrated in Fig.~\ref{fig:exp_main}. When analyzing the generated sequential frames, we specifically focus on the interaction and mutual engagement between the two participants. Because existing methods lack explicit interaction modeling, they typically struggle to capture co-motion dynamics, often producing static, independent subjects when rendering two agents within the same scene. In contrast, driven by our Cross-Modality Visual Guidance, our framework consistently generates natural, contextually grounded interactions. For instance, as shown on the left of Fig.~\ref{fig:exp_main}, only our method enables the listener to turn his head toward the active speaker. Similarly, on the right, our framework uniquely produces natural side-profile nodding in response to the speaker's cues. Furthermore, to intuitively visualize this dynamic engagement, we provide accumulated motion heatmaps in the bottom row of Fig.~\ref{fig:exp_main}. These heatmaps track the overall motion intensity of the characters over time, clearly demonstrating that the listener's reaction intensity in InterDyad is substantially stronger and more active compared to the other baselines. Overall, InterDyad ensures that listener responses are both frequent and triggered at the appropriate semantic moments, significantly enhancing the expressiveness of the dyadic interaction while preserving precise lip synchronization during active speech. Additional dynamic results are provided in the supplementary demo video.

\subsection{Ablation Study}
To thoroughly validate the effectiveness of our framework, we conduct ablation studies focusing on the architectural designs for dyadic interaction and the proposed inference optimization strategy.

\textbf{Architectures for Dyadic Interaction.} We evaluate the core components of our multimodal interaction module using the proposed DI-Sync and DI-Sali metrics, as showed in Tab.~\ref{tab:ablation}. First, when we remove the Interactivity Injection Module during the inference stage, the framework loses its direct control over behaviors. It reduces the overall sense of interaction, making the communication between the two agents feel less harmonious and results in a noticeable drop in both DI-Sync and DI-Sali scores. Second, to validate the necessity of MLLM-driven semantic reasoning, we replace our Modality Alignment module with a simple Audio-to-Motion MLP encoder. While this standard network can capture basic audio features, it struggles to understand the deeper context of the conversation and leading to a decline in the DI-Sync and DI-Sali metrics.

\begin{wraptable}{r}{0.5\textwidth}
  \centering
  \caption{Ablation study on the core architectural designs. Best results are \textbf{bolded}.}
  \label{tab:ablation}
  \resizebox{1\linewidth}{!}{
  \begin{tabular}{c cc}
        \toprule
        \multirow{2}{*}{Configuration} & \multicolumn{2}{c}{Interactivity (Ours)} \\
        \cmidrule(lr){2-3}
        
        & DI-Sync $\uparrow$ & DI-Sali $\uparrow$ \\
        \midrule
        
        w/o Interactivity Injection & 0.2041 & 0.7500 \\ 
        w/o MLLM Modality Alignment    & 0.2220 & 0.7800 \\ 
        \midrule
        
        \textbf{Ours}  & \textbf{0.2747} & \textbf{1.2349} \\
        \bottomrule
      \end{tabular}
  }
\end{wraptable}

\begin{figure}
    \centering
    \includegraphics[width=1.0\linewidth]{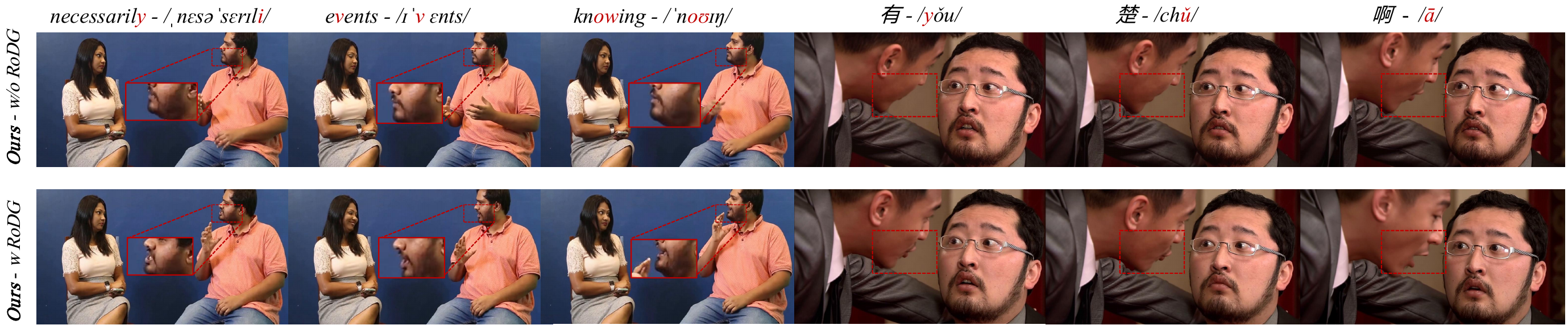}
    \caption{\textbf{Effect of RoDG on large-angle profile views.} Compared to the baseline without RoDG, our full method ensures robust and accurate lip reconstruction for the active speaker under severe perspective shifts.}
    \label{fig:ablation_fig}
    \vspace{-0.5cm}
\end{figure}

\textbf{Inference Optimization Strategy.} We further assess the impact of Role-aware Dyadic Gaussian Guidance (RoDG) on visual quality and audio-visual synchronization. In dyadic scenarios, extreme profile views often severely degrade lip-synchronization. By focusing the audio guidance directly on the active speaker's lip area using VAD, RoDG strengthens the lip generation. It is worth noting that standard quantitative metrics (e.g., Sync-C and Sync-D) are inherently less reliable for extreme side-profile views. Therefore, we conduct the lip-sync evaluation of these severe perspective shifts in a qualitative comparison manner. Specifically, we construct a specific hard-case set comprising 20 reference videos with challenging large-angle side profiles. Our results in Fig.~\ref{fig:ablation_fig} consistently exhibit accurate and robust lip movements under such extreme cases, which demonstrates the effectiveness of this training-free strategy. Please see the dynamic results provided in our supplementary demo video.

\section{Discussion and Conclusion}

\subsubsection{Limitation.}
While InterDyad enables explicit orchestration of high-fidelity dyadic interactions, some limitations remain. First, the framework is currently optimized for two-person scenarios and does not yet support multi-party conversations involving complex turn-taking among \textit{three or more participants}. 
We leave the extension to multi-person settings for future work.

\subsubsection{Conclusion.}
In this paper, we present InterDyad, a novel dyadic-conversational generation framework that bridges the gap between dual-track audio and realistic dyadic interaction patterns in shared-canvas scenes. By introducing the Interactivity Injector and MetaQuery-based Modality Alignment, our approach transcends traditional implicit generation, enabling the explicit orchestration of context-aware non-verbal behaviors from either reference videos or conversational audio. To ensure high-fidelity synthesis under extreme poses, we further propose RoDG to adaptively optimize lip-synchronization and coordination. Furthermore, we establish a dedicated evaluation suite with DI-Sync and DI-Sali metrics to quantify dyadic interactivity. Extensive experiments demonstrate that InterDyad significantly advances the state-of-the-art in synthesizing coherent, synchronized, and contextually grounded two-person interactions, paving the way for more immersive multi-person digital human experiences.

\clearpage  

\bibliographystyle{splncs04}
\bibliography{main}

\title{InterDyad: Interactive Dyadic Speech-to-Video Generation by Querying Intermediate Visual Guidance \\ {\large Supplementary Material}} 
\author{}
\institute{}
\maketitle

\section{Data Curation}
\label{supp:dataset}

We curate a large-scale, high-fidelity dyadic interaction dataset from extensive in-the-wild video repositories. The following sections detail our cascaded pipeline—spanning heuristic filtering, dyadic identification, and motion-aware optimization—designed to ensure both data quality and rich inter-agent dynamics.

\subsubsection{Heuristic Quality Filtering.}
To ensure the high fidelity of the source material, we implement a multi-stage filtering protocol. 
Firstly, we filter for videos with a resolution of at least 720P and a frame rate of at least 25 FPS. To maintain temporal consistency, we retain clips with a duration $T \in [T_{min}, T_{max}]$, where $T_{min} = 3\text{s}$ and $T_{max} = 10\text{s}$; sequences exceeding this range are truncated to the initial $10\text{s}$ window.
For camera stability, we compute the global motion vector and reject clips where the camera jitter score $s_{jitter} < \tau_{s}$ or the average optical flow $\mathbf{V}_{opt} > \tau_{v}$. 
For video quality, we enforce a clarity constraint based on a frame-wise scoring mechanism, discarding sequences where the mean clarity $\bar{C} < \tau_{c}$. 
This stage ensures that the input to our framework is free from severe motion blur and acquisition noise. 

\subsubsection{Dyadic Interaction Identification.}
We isolate dyadic candidates through a rigorous spatial-consistency check with DWPose detection~\cite{yang2023effective}. A video is classified as a "dyadic interaction" only if it satisfies the following criteria: \textbf{1)} Detection Co-occurrence: Exactly two body bounding boxes must be detected in at least $R_{body}$ of the total frames, and two facial bounding boxes in at least $R_{face}$ of the frames. \textbf{2)} Semantic Constraints: We employ a body-type classifier to filter for specific subjects (e.g., upper-body or full-body portraits), ensuring the interaction is centered on human agents rather than cluttered backgrounds. To ensure rich non-verbal expressiveness, we utilize DWPose to calculate the mean velocity of body keypoints, retaining only clips where the maximum velocity of hands or head $\mathbf{V}_{motion}$ exceeds $\tau_{m}$. This step effectively filters out static "talking head" sequences, ensuring the dataset contains high-quality interactive gestures and dynamic postures.

\subsubsection{Audio-Visual Alignment and Dynamic-Aware Segment Extraction.}
Following the pipeline of the MIT dataset~\cite{zhu2025multi} , we employ WhisperV~\cite{whispervideo} to segment raw videos into individual shots and track facial trajectories to ensure temporal consistency. Subsequently, we use TalkNet~\cite{beliaev2020talknet} to extract speaker-specific scores, which serve as spatial-temporal anchors to precisely align acoustic features with visual bounding boxes for each individual.

The final dataset comprises approximately $700,000$ high-quality clips, all standardized to a minimum resolution of $720$P at $25$ FPS, with durations ranging from $3$ to $10$ seconds.

\begin{table}[ht]
\centering
\caption{Summary of Notations and Thresholds for Dataset Curation.}
\label{tab:dataset_notations}
\small
\begin{tabular}{ccc}
\toprule
\textbf{Symbol} & \textbf{Description} & \textbf{Value/Standard} \\ \midrule
$\tau_{s}$ & Minimum threshold for camera stability & 3.66 \\
$\tau_{v}$ & Maximum threshold for camera motion/flow & 6.0 \\
$\bar{C}$ & Mean clarity score of a video sequence & - \\
$\tau_{c}$ & Minimum threshold for visual clarity & 0.95 \\ \midrule
$R_{body}$ & Frame ratio requirement for dual body detection & 80\% \\
$R_{face}$ & Frame ratio requirement for dual face detection & 30\% \\ 
$\tau_{m}$ & Threshold for velocity of hands or head & 0.12 \\

\bottomrule
\end{tabular}
\end{table}

\section{Supplementary Experiments}
\label{supp:exp}

\subsection{Mono Speech-to-Video Generation}

\begin{figure*}[!t]
\centering
\includegraphics[width=\linewidth]{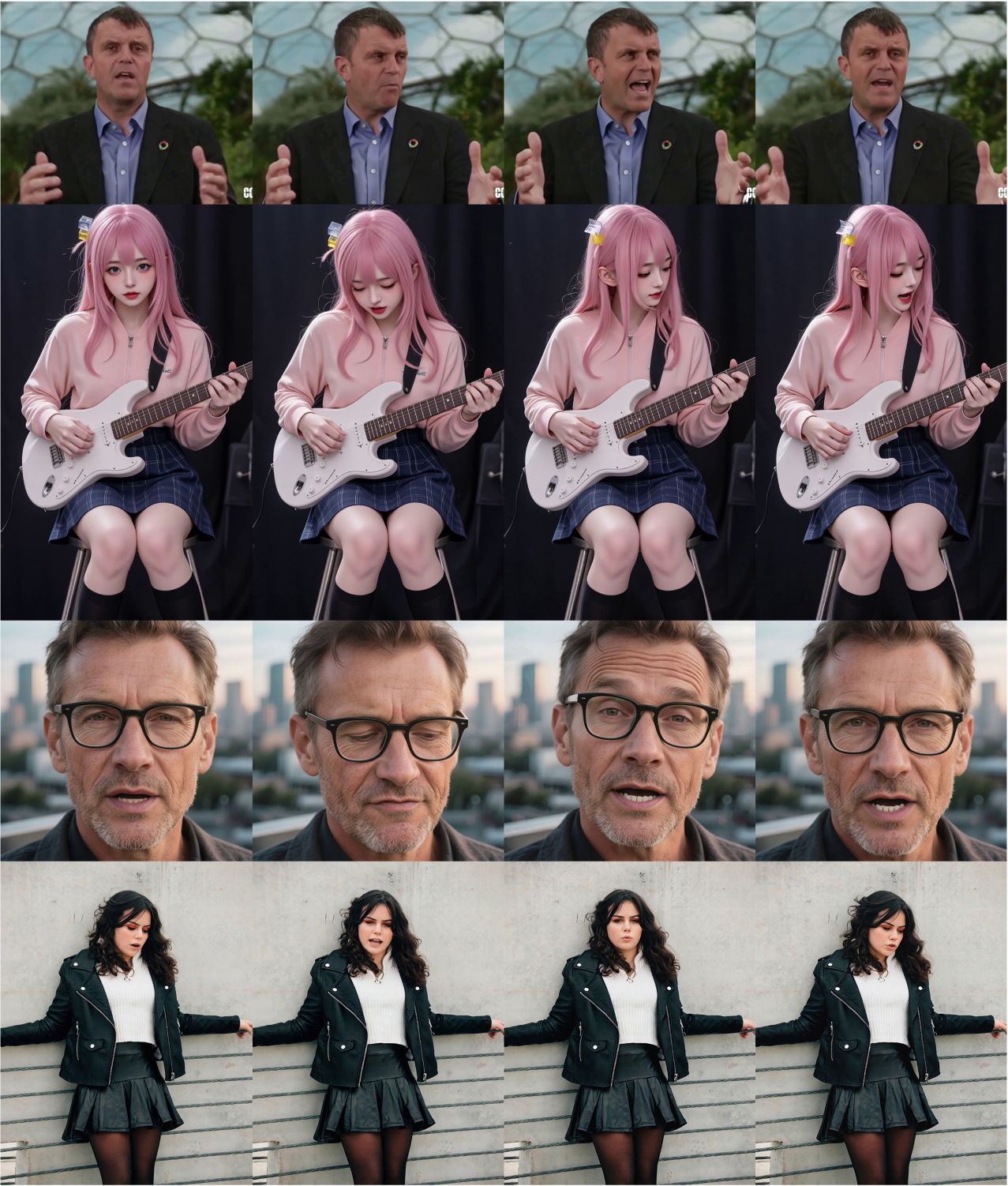}
\vspace{-0.25cm}
\caption{
\textbf{Qualitative results of our method on the single-person dataset.}
Each row represents a distinct conversation case, while columns show selected frames at different time steps to illustrate temporal progression. Our model consistently generates high-fidelity facial details and natural head motions that are precisely synchronized with the input audio across various subjects and expressive states.
}
\label{fig:single_case}
\vspace{-0.25cm}
\end{figure*}

\begin{table}[t]
  \centering
  \caption{Quantitative comparison on our single-person dataset. LongCat-VA denotes LongCat-Video-Avatar. ``$\uparrow$'' indicates higher is better, and ``$\downarrow$'' indicates lower is better. Best results are \textbf{bolded}, and the second best are \underline{underlined}.}
  \label{tab:single_result}
  \resizebox{0.85\linewidth}{!}{
  \begin{tabular}{c ccc cc}
    \toprule
    \multirow{2}{*}{Method} & \multicolumn{3}{c}{Visual \& Identity} & \multicolumn{2}{c}{Synchronization} \\
    \cmidrule(lr){2-4} \cmidrule(lr){5-6}
    & FID $\downarrow$ & FVD $\downarrow$ & ID-Cons $\uparrow$ & Sync-C $\uparrow$ & Sync-D $\downarrow$ \\
    \midrule
    MultiTalk \cite{multitalk}       & 34.3847 & 345.4332 & 0.6811 & \underline{8.5122} & 2.1538 \\
    InfiniteTalk \cite{infinitetalk} & \underline{33.1035} & \textbf{329.2524} & \textbf{0.8260} & 8.5060 & 1.7574 \\
    LongCat-VA \cite{longcat}        & 49.1042 & 491.3213 & 0.7332 & 7.7535 & \underline{1.5684} \\
    Wan-S2V \cite{wans2v}        & 57.8167 & 585.7959 & 0.7093 & 7.3267 & 6.0978 \\
    \midrule
    \textbf{Ours}                    & \textbf{33.0428} & \underline{340.4103} & \underline{0.7916} & \textbf{8.5386} & \textbf{1.4962} \\
    \bottomrule
  \end{tabular}
  }
\end{table}

\noindent \textbf{Test Datasets and Metrics.} To evaluate the generalization of our method, we construct a test set of \textbf{250 cases} by randomly sampling from four benchmark datasets: \textbf{HDTF}~\cite{zhang2021flow}, \textbf{CelebV-HQ}~\cite{zhu2022celebv}, \textbf{EMTD}~\cite{rang2025emtd}, and the \textbf{Soul}~\cite{soul2025} dataset. This combination covers a wide spectrum of facial dynamics, body movements, and expressive talking styles. Consistent with the main paper, we evaluate visual fidelity using Fréchet Video Distance (FVD)~\cite{unterthiner2018towards} and Fréchet Inception Distance (FID)~\cite{heusel2017gans}, measure lip-sync accuracy via Sync-C and Sync-D scores, and verify ID-Cons  by computing ArcFace~\cite{Deng_2022} similarity between generated and reference facial crops. For comparison, we adopt the baseline methods used in the main paper, including MultiTalk~\cite{multitalk}, InfiniteTalk~\cite{infinitetalk}, and LongCat-VA~\cite{longcat}. Furthermore, we extend our comparison to include Wan-S2V~\cite{wans2v}, a state-of-the-art mono S2V framework.

\noindent \textbf{Quantitative and Qualitative Analysis.} As shown in Tab.~\ref{tab:single_result}, our approach achieves highly competitive performance across all dimensions. In terms of spatial visual quality, our method attains the best FID score, indicating superior frame-level synthesis. Our model consistently ranks among the top performers, demonstrating its effectiveness for high-fidelity talking head generation.

As evidenced by the qualitative examples in Fig.~\ref{fig:single_case}, our method generates sharp facial textures and maintains consistent identity features. Although InfiniteTalk achieves slightly higher ID-Cons and a better FVD score, our method remains a strong contender, ranking second in both metrics. This performance gap can be attributed to our framework's prioritization of motion expressiveness; as illustrated by the temporal sequences in Fig.~\ref{fig:single_case}, our model produces vivid, natural head dynamics, whereas baselines with more conservative motions often benefit from higher temporal stability at the cost of realism.

Furthermore, our framework excels in audio-visual alignment, achieving state-of-the-art results in both Sync-C and Sync-D. The temporal frame sequences in Fig.~\ref{fig:single_case} demonstrate that our model produces precise and diverse lip shapes accurately synchronized with the driving audio. This confirms our method’s capability to translate complex audio signals into realistic movements while maintaining naturalness. Additional dynamic results are provided on the supplementary project page.

\section{Ethical Considerations}
\label{supp:ethical}

Our research aims to enhance the realism of audio-driven dyadic interaction synthesis. While recognizing the potential of this technology, we also acknowledge its dual-use risks and hereby state our ethical stance:

\textbf{Prevention of Misuse and Impersonation}: Similar to other human animation methodologies, this framework carries a risk of malicious exploitation. As the technology can model natural, reactive interpersonal dynamics, synthetic content may become increasingly difficult to distinguish from authentic footage. We strictly oppose any use of this framework to generate or manipulate an individual's likeness without their explicit consent.

\textbf{Preservation of Social Trust}: The proliferation of high-fidelity synthetic media challenges the authenticity of digital identity. If used to fabricate conversations or collaborative scenarios that never occurred, such technology could cause reputational harm and undermine public trust in visual media.

\textbf{Commitment to Responsible Research}: This work is intended to advance creative industries, education, and accessibility. We support the development of robust content detection technologies to mitigate potential abuse. We urge all users to adhere to professional ethical standards, respect privacy and portrait rights, and strictly avoid applying this technology for deceptive or misleading purposes.

\end{document}